# Identification via Retinal Vessels Combining LBP and HOG

**Ali Noori**[†]
*Alinoori@outlook.com*
Islamic Azad University of Mashhad, Iran

**Abstract**
With development of information technology and necessity for high security, using different identification methods has become very important. Each biometric feature has its own advantages and disadvantages and choosing each of them depends on our usage. Retinal scanning is a bio scale method for identification. The retina is composed of vessels and optical disk. The vessels distribution pattern is one the remarkable retinal identification methods. In this paper, a new approach is presented for identification via retinal images using LBP and hog methods. In the proposed method, it will be tried to separate the retinal vessels accurately via machine vision techniques which will have good sustainability in rotation and size change. HOG-based or LBP-based methods or their combination can be used for separation and also HSV color space can be used too. Having extracted the features, the similarity criteria can be used for identification. The implementation of proposed method and its comparison with one of the newly-presented methods in this area shows better performance of the proposed method.

## 1. Introduction

Recently a technology has been raised for identification which is based on a biometric field. Biometric access control provide automatic methods for identification based on physical features such as fingerprint, hand geometry, face, iris, retina or several behavioral features such as talking, handwriting, or keystroke patterns. Among different biometric methods, physical features are more sustainable than behavioral features since physical features are fixed unless serious harms leads to their loss while behavioral features may vary in anxiety, fatigue or illness [1].

Each biometric feature has its own weak and strong points and choosing each of them depends on our usage. Retinal scanning is a bio scale method for identification. The retina is composed of vessels and optical disk. The vessels distribution pattern is one the remarkable retinal identification methods. The information related to thickness, length, curvature, and manner of distributing blood vessels are very important in images. Retinal Identification provides high level of security since the pattern of blood vessels has many uniformities and stability in life-long.

As regards the development of information technology and the necessity for high security, different identification methods have become very important. An identification system should have zero error, low cost, high speed and non-aggression. As we know, an identification system acts based on a unique feature. The stability and no general changes of this feature guarantees the efficiency of the system. No much stability is seen in identifying face and voice. In retina, it is not the same. The pattern of vessels distribution is fixed along the life and it has high precision in retinal-based identification. In retinal images, the appearance of vessels, i.e., their distribution pattern, is the most remarkable index and whereas the vessels shapes are suitable tool in presenting features while automatic identification, their detection will be suitable in identification. The vessels detection problem is that in usual retinal images, there is no much difference between vessels and background and using usual methods such as Gradient will not be working. One of the problems of previous methods in retinal identification can be low precision in identification for diabetic patients, low precision in images analysis for different brightness, impossibility of extracting non-major retinal vessels via image-segmentation methods, and low precision in gradient-based methods. In this research, it is tried to provide a new approach for analyzing retinal images to solve the problems of previous methods. One of the methods which can be used in this regard are modeling vessels lines in retina. The human body features such as fingerprint, face, palm, and iris were used for identification. Identification via retinal vessels can be used as one of the strongest identification systems. One of the advantages of retina is that it is fixed in life long while in other methods, the variability is possible via several methods such as plastic surgery [20]. In the proposed method, it will be tried to separate the retinal vessels via machine vision techniques to have good sustainability in rotation and size change and HOG-based or LBP-based or their combination methods can be used for separation and also HSV color space can be used too. Having extracted the feature, similarity can be used for identification.

In continuing, part 2 will study the previous presented methods in retinal identification. Part 3 will express the method based on the combination of lbp and hog techniques. In part 4, the proposed method will be evaluated and its function will be studies and finally part 5 will give conclusion and summing up.

## 2. Review of Literature

Retinal identification has been considered in many researches, some of which will be discussed in continuing. Farzin et.al (2008) presented a new retinal identification system. In that, a new method based on features obtained from retinal images was proposed. This method is composed of three major modules including division of vessels, producing features and adjusting features. The module of vessels dividing results in the pattern of retinal vessels from images. The production module includes optical disk detection and choosing circular zone around optical disk from divided images and then a fixed rotation mold is created via a polar transformation. At next stage, this mold is transferred to separate vessels via wavelet in three different scales which are studied due to their diameter size.

Diseases such as diabetes, hypertension, Coronary Artery Sclerosis and Vascular Sclerosis influence on retina. For example, diabetes is revealed as bright spots in retina which misdiagnose optical disk. Optical disk in high brightness images and its bigger dimensions influence on other low brightness and small images such as vessels and their effect is not evaluated well. This feature detection method in a small database can obtain a high percent recognition while in bigger database, the recognition authenticity drops down since the pattern of vessels distribution is fixed along life and it is unique for different people. Therefore, it is not logical to use all parts of retinal images in feature detection. Guo et.al (2005) provided a system for vessels detection. They used the distribution of vessels and omission of other parts of retina like optical disk for feature detection and used the curvature vector of skeleton of vessels for differentiating different images.

Tabatabaee et.al (2006) presented a new algorithm based of fuzzy-clustering algorithm. They used roaring wave model and snakes for finding optical disk location. Shahnaz et.al (2007) presented a new method based on wavelet energy feature (WEF) for multi-decision analysis. WEF can reflect the distribution of waves energy with different thicknesses and width in different waves analysis levels.

Oinonen et.al (2010) presented a new method for identification based of auxiliary features. The proposed method includes three steps: vessels segmentation, extracting features, adjusting features. The vessels segmentation can be pre-process for features detection phase.

Dehghani et.al (2013) detects human based on retinal images using new similarity function. This paper has presented a new method for identification based on extracted features from retinal images. In the presented method, the feature extraction methods such as phase correlation method and corresponded features for detection were used.

Lajevardi (2013) presented a retinal security system based on chart matching biometric. This paper focused on automatic network and presented a framework based on BGM. The retinal vessels are extracted via similar filters in frequency and morphological operators. Meanwhile, Ikibas et.al (2013) presented a human detection system via retinal vessels in images.

## 3. Proposed Method

In this chapter, we will study the proposed method for retinal diagnosis in identification. At first, we describe the composing parts of proposed method in the flowchart of desired method which displays the general structural as conceptually. Fig. 1 shows the flowchart of presented method in this research.

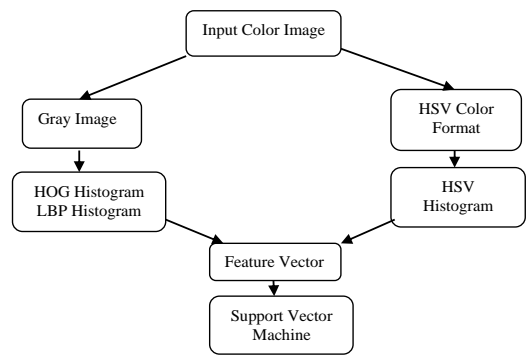

**Fig. 1**: Flowchart of Proposed Method to Solve Identification Problem based on Retinal Images

We need proper color format to extract color information which facilitates the image segmentation. There are different color formats, each of which have different features. In this regard, HSV color format has good properties for image segmentation. This format has three matrices; the first matrix is Hue and its values are related to inherent element of color. In this research, the Hue values in 10 columns in 32*32 were computed in 32-distance points which have correlation to classify the histogram retinal image and they are use with image texture information in LBP (Local Binary Pattern) and HOG (Histogram of Oriented Gradient) to make feature vector. Fig. 2 displays the main image used in RGB (Red Blue Green) format in this research and its HSV format.

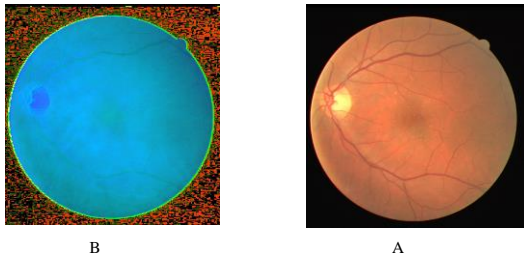

**Fig. 2** : A, Color Format Image RGB of retina and B, its HSV format

There are many methods to extract the features of tissue and among them, local binary pattern in its main and modified form has been considered a lot by the specialists due to its simplicity in implementation and extraction of proper features with high classification precision. In this research, we use its main form and extract the available binary patterns via radium neighborhood unit in image and in 32*32 range, we create histogram of that value in [0-255] and the resulted histogram is used in making feature vector after normalization to be used in tissue feature. Fig. 3 displays the neighborhood radius and number of used neighbors in this method due to different values of the two mentioned parameters.

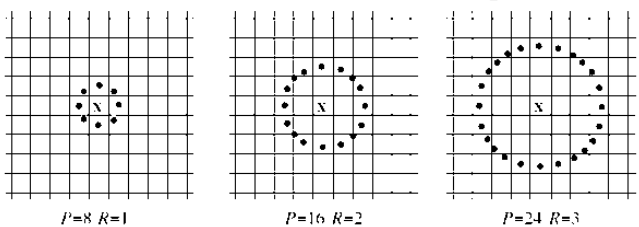

**Fig. 3:** R neighborhood radius and P number of used neighbors in LBP method for different values

In this paper, HOG method is used for extract features. This method is one of the most used methods of feature vector extraction and is used successfully in different applications such as pedestrian detection. The first step is extraction of feature and this feature extraction includes computing horizontal and vertical axis of gradients and also computing the size and direction of gradients. The below image is a 64*128 image. We segment this image to the displayed cells. We classify these cells and each 4 cells are called a block. The number of cells in each block is 2*2. On the other hand, each cell is 8*8 size. Therefore, the blocks are 16*16. So, we have 7 blocks in horizontal side and 15 blocks in vertical side. So generally, in each image, we will have 7*15 blocks. For optimizing the resolution, we consider the blocks such that each block has 50% overlap with neighbor block. This technique is used due to its high precision. For example, if the human image is exactly put between two blocks, it will not be precise estimation, while we can have information from previous block with this technique and have estimation that is more precise. Finally, we extract oriented gradients in each histogram block and reach to a general histogram on image upon annexing all histograms. Fig. 4 displays the more detailed manner of feature extraction with this method.

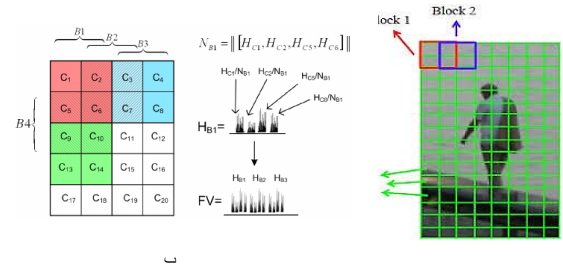

**Fig. 4:** Schematic Manner of Extracting histogram of oriented gradients or HOG

Having extracted the features, it is the turn for prediction and final identification. In this paper, SVM classification is used in this regard. It can be said that support vector machine is one of the best classification methods which is used in most machine vision application for classifying feature vectors. Since this classification method has many advantages, it has been used as classifier of feature vectors in this research. To classify support vector machine (SVM), audit function may be used too [35]:

$$f(x) = \sum_{i=1}^{l} y_i \alpha_i k(x, x_i) + b$$

Where l is number of training samples, b is bias 1 and xi, yi are ith training sample and its label. Yi labels are use in 1 or -1 which specifies the class of related samples.  function is a kernel function which should be applied in Mercer.

Kernel in [35] is defined as two Gaussian and polynomial forms.
.

Polynomial kernel: $k(x, x') = (1 + x.x')^p$

RBF Kernel: $k(x, x') = \exp(-\frac{\|x-x'\|^2}{2\sigma^2})$

Where $p$ and $\sigma$ are corresponded kernel parameters.

$\alpha_i$ coefficient (i=1,…,l) is defined as equation (7-2) upon solving the following optimization problem.

$$L(\alpha) = \sum_{i=1}^{l} \alpha_i - \frac{1}{2}\sum_{i,j}^{l} \alpha_i \alpha_j y_i y_j k(x_i, x_j),$$

$$0 \leq \alpha_i \leq C, i = 1, \dots, l$$

$$and \sum_{i=1}^{l} \alpha_i y_i = 0$$

Condition:

Where C is a classifier parameter. Many methods were proposed for solving optimization problem especially when there are big training samples among them.

The high diagnosis percent depends on choosing type of core and number of features. For example, if the used main feature is merely computed in two horizontal and vertical sides, the diagnosis percent reduces a little [34, 35].

## 4. Evaluation of Proposed Method

In this chapter, we are dealing with presenting some results to evaluate the proposed method in previous chapter. To evaluate the proposed method, DRIVE images bank was used. This images bank includes 40 color images of retina and vessels from 40 different persons. In this research, we used color retinal images for identification. In this regard, at first contrast of images was optimized and feature vectors of retinal color images were extracted in HSV format via histogram and LBP and HOG methods. 40 feature vectors of these images were extracted and were considered as training set.

To establish test set, the retinal images were changed randomly without preprocessing, optimizing contrast and a little rotation in [-10, 10] gradient and feature vectors were extracted and are considered as test set. Training set is used to train linear support vector machine and test set is used for evaluation. Fig. 5 shows a sample of available images in used images bank.

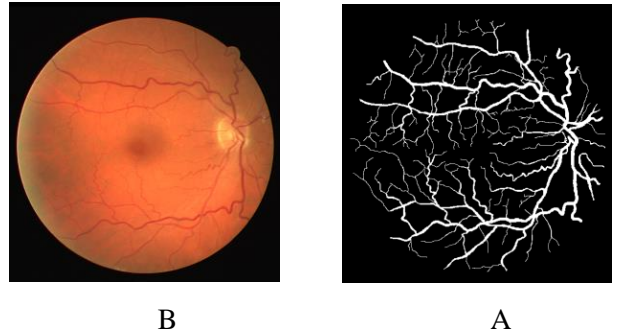

B  A

**Fig. 5:** Image A, segmented vessels and B color image related to retina

As it was told, we used feature vectors classification with support vector machine for classification and final identification in this paper. Classifying feature vectors is a very challenging problem because there is only one image for each person to train classifier while support vector machine shows have better performance and the obtained results are acceptable. To prove the efficiency of proposed method in [33], we evaluate the images used in this research and presented the obtained results in table 1. The presented method in this reference was based on SIFT (Scale Invariant Feature Transform) which is somehow sustainable in gradient and dimensions of input image. Having adjusted these features, the retinal images belonged to similar persons are identifiable with high precision. Fig. 6 shows a sample of adjusting retinal images via these features.

For better evaluation, we study the proposed method for different evaluation data. for this end, we repeat the evaluation process 10 times and presented the mean of identification precision for different data in table 1.

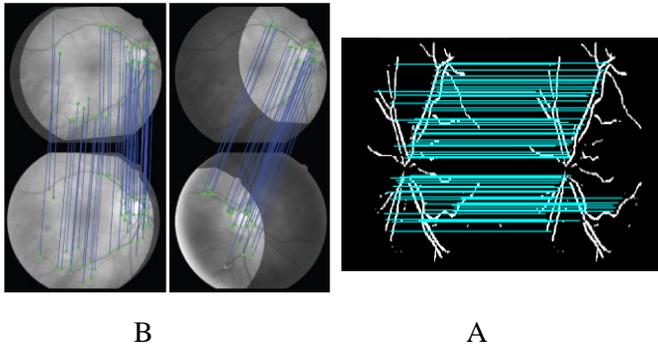

B                           A

**Fig. 6:** Retinal diagnosis upon adjusting SIFT features

**Table 1:** The obtained results from classifying feature vectors with proposed method in comparison with [33]

| Type of Data | Number of Images | Precision of proposed Method | Precision of reference method [33] |
|---|---|---|---|
| Training Data | 40 | 100% | 100% |
| Evaluation Data | 40 | 98.5% | 98% |

The results of proposed method are acceptable and promising due to less number of images; one image was belonged to each person. Meanwhile the proposed method shows better results in similar condition compared to reference method [33].

## 5- Conclusion

Nowadays identification has become very important. Different methods have been proposed for identification varying from fingerprint to iris and recently retina. In recent years, many methods have been presented for identification via retinal images; this research aims at improving the available methods for identification based on retinal images. The presented method is based on HOG, LBP methods and histogram of Hue matrix values in HSC color format. The feature vectors were extracted from 40 images, one for each person, and test vector used 40 changed images to evaluate via support vector machine. 40 training vectors were classified to train support vector machine and 40 test vectors were classified to evaluate via support vector machine. The process of changing images and extracting test vectors were repeated for 10 times and the results of classification were obtained via mean. To evaluate the proposed method, a comparison was done with reference method [33] where the mean of precision is 98%, the mean of precision in the proposed method is 98.5% and the results are promising while the dimensions of feature vector for proposed method is high. To improve the performance of proposed method, we can use different methods of choosing feature vector which may reduce the dimensions of feature vector and can increase the diagnosis precision of the desired method. In future works, the adjusted approach of local histograms similar to which was used in [33] can be used to make big feature vector with local feature histograms to present the advantages of both desired method in a novel approach.